\documentclass[11pt]{article}

\usepackage[final]{acl}

\usepackage{times}
\usepackage{latexsym}

\usepackage[T1]{fontenc}

\usepackage[utf8]{inputenc}

\usepackage{microtype}
\usepackage{amsmath}
\usepackage{comment}
\usepackage{booktabs}
\usepackage{CJKutf8}    
\newcommand{\tablesize}{\small}

\usepackage{inconsolata}

\usepackage{algorithm}
\usepackage{algpseudocode}

\usepackage{graphicx}

\usepackage{environ}   

\algrenewcommand\algorithmicfunction{\textbf{Algorithm}}

\algtext*{EndIf}
\algtext*{EndFor}
\algtext*{EndFunction}
\algtext*{EndWhile}

\newcommand{\task}{sense generation}

\newcommand{\exnet}{\texttt{ExpandNet}}
\newcommand{\dbalign}{\texttt{DBAlign}}

\newcommand{\corpustwenty}{\texttt{SC20}} \newcommand{\corpusfull}{\texttt{SC}}
\newcommand{\uwn}{\texttt{UWN}}
\newcommand{\dictbaseline}{\texttt{Dict}}
\newcommand{\oconefour}{\texttt{OC14}}
\newcommand{\mtwothree}{\texttt{M23}}
\newcommand{\llmbase}{\texttt{ChatGPT}}

\title{Generating Concept Lexicalizations via \\
Dictionary-Based Cross-Lingual Sense Projection}

\author{
David Basil, 
Chirooth Girigowda, 
Bradley Hauer, \\
{\bf 
Sahir Momin,  
Ning Shi,
Grzegorz Kondrak
}
\\
  Alberta Machine Intelligence Institute \\
  Department of Computing Science \\ 
  University of Alberta, Edmonton, Canada \\
  \texttt{{\{dbasil1,gkondrak\}@ualberta.ca}} \\
}

\begin{document}
\maketitle

\providecommand{\keywords}{}
\RenewEnviron{keywords}[1]{}

\providecommand{\copyrightnotice}{}
\renewcommand{\copyrightnotice}{}

\providecommand{\printbibliography}{}
\renewcommand{\printbibliography}[1][]{\bibliography{references}}

\begin{abstract}
We study the task of automatically expanding WordNet-style lexical resources to new languages 
through \task{}.
We generate senses by associating target-language lemmas with existing lexical concepts
via semantic projection.
Given a sense-tagged English corpus
and its translation,
our method projects the annotated synsets 
onto aligned target-language tokens 
and assigns the corresponding lemmas 
to those synsets.
To generate alignments 
and ensure their quality,
we augment a pretrained base aligner 
with a bilingual dictionary, 
which is also used to
filter incorrect sense projections.
We evaluate the method on multiple languages,
comparing it to prior methods,
as well as dictionary-based
and  large language model baselines.
Results show that the proposed project-and-filter
strategy improves precision while remaining interpretable and resource-efficient.
We release our code, documentation, and generated sense 
inventories at \url{https://github.com/UAlberta-NLP/ExpandNet}.
\end{abstract}

\copyrightnotice

\section{Introduction}
\label{intro}

A wordnet is 
a semantic resource that organizes words into synonym sets,
or \emph{synsets}, providing a structured representation of word meaning. 
Each synset represents a distinct lexical concept 
and contains the lemmas that express it. 
Wordnets are widely used
throughout natural language processing (NLP).
In word sense disambiguation (WSD),
they serve as sense inventories,
enabling the mapping of words in context to structured lexical knowledge \cite{pasini2021xlwsd}.
They also support semantic parsing \cite{navigli2022}, machine translation evaluation \cite{campolungo2022}, and the word-in-context task \cite{pilehvar2019}. Despite their importance, 
automating wordnet construction remains challenging,
particularly in the multilingual setting, where words from different languages must be 
organized into a shared semantic ontology.
A central component of this process is {\task}.
Because a word sense pairs a lemma with a synset,
\emph{generating} a sense is equivalent to adding new lemmas to existing synsets.

Prior work has leveraged parallel data to generate senses.
\citet{oliver2012a,oliver2012b,oliver-climent-2014-automatic} 
were the first to explicitly
project English sense annotations onto lexical translations
to extend wordnets.
More recently, \citet{martelli-etal-2023-lexicomatic} combined WSD, translation, and alignment 
on an in-house English corpus 
to construct multilingual semantic dictionaries.
While effective, these
projection-based methods 
typically rely on corpus-level statistics,
making them sensitive to corpus size and distribution, which limits their scalability.

To address this limitation,
we propose a finer-grained,
sentence-level method for generating target language senses
through dictionary-guided projection.
Building on the semantic projection framework,
we derive these new senses from translations of a sense-annotated source text. 
To ensure high-quality word alignment, 
we augment a pretrained aligner with a bilingual dictionary.
We then use this bilingual dictionary to filter semantic projections, significantly reducing erroneous synset assignments.
To our knowledge,
this is the first application of dictionary-based filtering to \task{}.
By functioning strictly at the sentence level,
our algorithm eliminates the reliance on corpus-level
statistics. Ultimately, 
we envision our system as a high-precision recommendation tool,
generating candidate senses reliable enough for 
rapid verification by native speakers.

Our project-and-filter procedure rests on three core assumptions 
about word senses and translation.
First,
we assume that each content word
in context
expresses a single
lexical concept,
allowing our system to eliminate
competing synsets rather
than retaining
multiple candidates for a single token.
Second, in a multilingual wordnet,
synonymy functions
as a test of conceptual
identity:
lemmas that
are synonymous
realize the same concept.
Third, when the alignment of words in a sentence and its translation 
is supported by lexical evidence
(e.g., a bilingual dictionary or a clear cognate relationship),
it represents a literal translation. 
The aligned words therefore express the same concept,
allowing
the target lemma to inherit
the synset
of the disambiguated source token.

\begin{figure}[t]
    \centering
    \includegraphics[width=0.9\columnwidth]{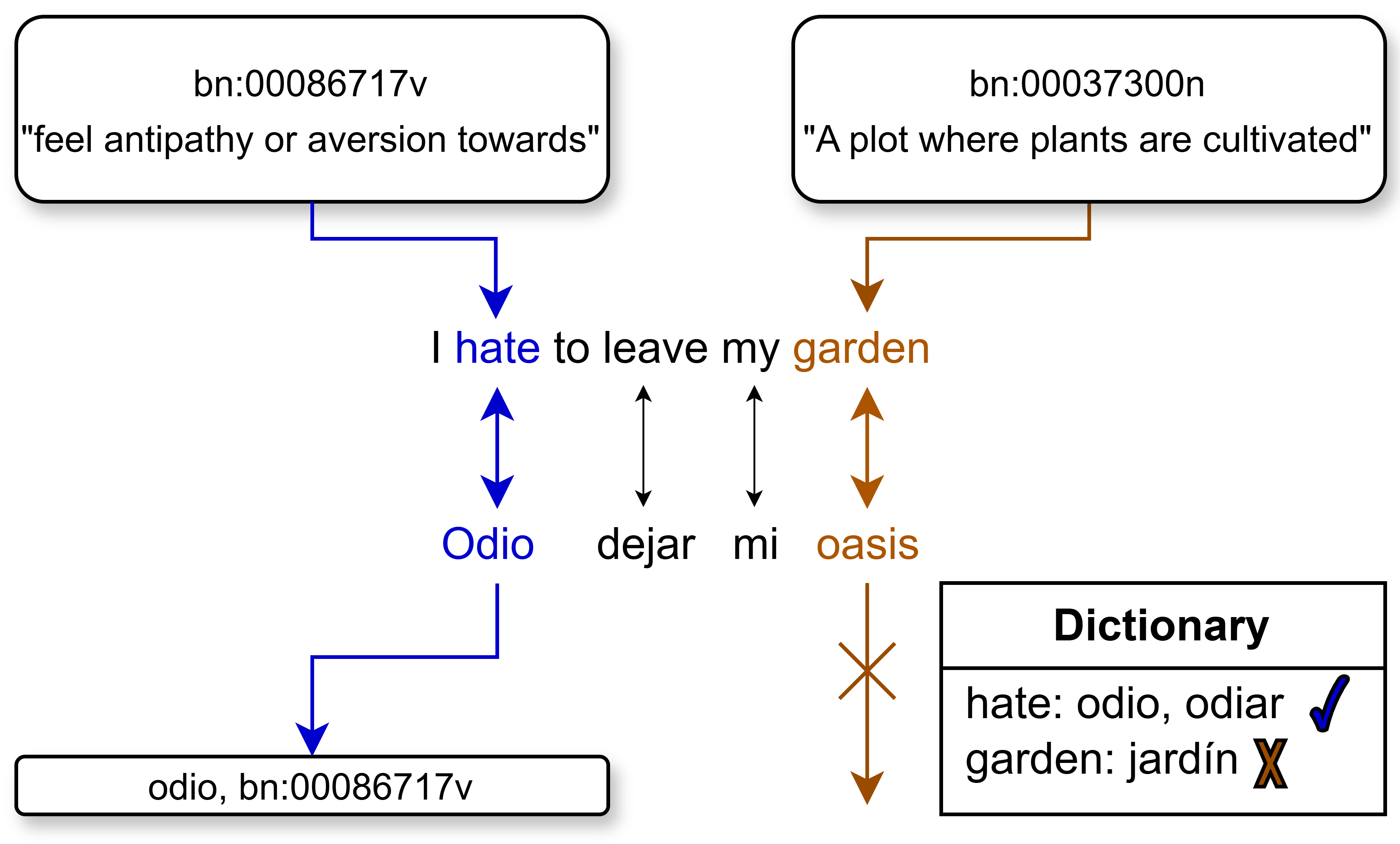}
    \caption{An illustration of our project-and-filter algorithm.}
    \label{fig:expandnet}
\end{figure}

Figure~\ref{fig:expandnet} illustrates how dictionary filtering governs sense projection in our algorithm.
The source sentence includes two sense-tagged words,  
with their corresponding translations identified through word alignment. 
The first aligned pair is confirmed by the bilingual dictionary, 
allowing the sense tag assigned to \emph{hate} to be projected onto \emph{odio}, creating the new sense (\emph{odio}, \texttt{bn:00086717v}).
In contrast, 
because the second aligned pair 
is not confirmed by
the dictionary,
the translation of \emph{garden} is 
considered non-literal; 
the alignment is discarded, 
and no sense is generated.

Our experiments across four
diverse language pairs demonstrate 
that our approach improves precision compared to prior projection-based 
methods.
Dictionary filtering drives this
precision by reducing
spurious sense assignments,
while our dictionary-augmented aligner 
improves coverage by recovering
links that would otherwise be missed.
Overall, the method achieves competitive performance
while remaining highly interpretable. Furthermore, our error analysis indicates that most remaining errors arise from
noisy resources or upstream mistakes in translation or alignment,
rather than from limitations of our methodology itself. 

Our core contributions are as follows:
\begin{enumerate}
                        \item We propose a robust, sentence-level method for \task{} that operates independently of corpus-level statistics.
                                    \item We introduce and validate a dictionary-augmented alignment method.
                \item We demonstrate the effectiveness of 
    dictionary filtering
    for projection-based wordnet expansion.
                \end{enumerate}

\section{Related Work}
\label{relwork}

The development of Princeton WordNet \cite{miller1990lexi} 
spurred the creation of wordnets in various languages,
many of which have been unified through the 
Open Multilingual Wordnet project
\cite{bond2012survey}.
Similarly, regional efforts such as IndoWordNet \cite{bhattacharyya-2010-indowordnet}
and Asian Wordnet \cite{sornlertlamvanich-etal-2010-language} seek to construct wordnets
for multiple languages.
However, because these resources rely heavily on manual curation,
they are often constrained by the high cost of human annotation,
resulting in a scarcity of high-quality,
widely multilingual wordnets. 

To address this scarcity, researchers have explored various automated approaches. 
One such task, word sense induction (WSI), is closely related to \task{},
though it operates from an opposing perspective.
While WSI induces senses from a text in a ``bottom-up'' fashion,
effectively discovering concepts through methods such as clustering \cite{lin-pantel-2002-concept, di-marco-navigli-2013-clustering},
\task{} is a ``top-down'' process.
It focuses on pairing words with pre-existing concepts within a structured ontology,
thereby populating a wordnet rather than learning its structure from scratch.

The earliest large-scale attempt
to  automatically expand a wordnet
was the \emph{Universal WordNet} (UWN) by \citet{demelo2009}.
Their method links target-language words to wordnet synsets
via a supervised classifier,
drawing on existing wordnets,
translation dictionaries, and parallel corpora.
Although their approach differs from ours
and relies on a broader set of resources,
their task is fundamentally the same. As UWN is freely available,
it provides a natural point of comparison
for evaluating our method.

BabelNet \cite{navigli2010} is a more comprehensive and actively-developed
multilingual wordnet.
It originated as an automatic linking of multiple online knowledge bases,
including Princeton WordNet \cite{miller1990lexi} and Wikipedia,
and has since been expanded with additional resources
and manual refinement.
BabelNet consists of multilingual synonym sets,
each corresponding to a concept
and including lexicalizations in multiple languages.
Due to its prominence and manual curation, 
we treat BabelNet as a gold standard rather than a competing system.

Prior work has
demonstrated that translations provide valuable
sense-specific information.
Classical
WSD 
methods
often used translations as sense inventories,
representing a word's
meaning by how it was translated \cite{gale1992a}.
Even after wordnets and dictionaries became the
standard for sense inventories,
translation-based WSD methods continued to be developed 
\citep{apidianaki-gong-2015-limsi,luan2020}.
In particular, translations 
have been used to efficiently sense-tag text at scale,
thereby creating training data for WSD 
\citep{pasini2021xlwsd},
an approach which remains popular as a way to address the knowledge acquisition bottleneck \cite{pasini2021}.

Projection leverages translations to 
transfer linguistic annotations from one language to another, usually a resource-poorer language.
\citet{yarowsky-etal-2001-inducing} argued that projection
provides a means to induce tools for part-of-speech tagging,
named entity recognition, and noun-phrase bracketing.
Since then, projection has been used for a variety of tasks,
including tree parsing \cite{akbik-vollgraf-2018-zap},
relation extraction \cite{faruqui-kumar-2015-multilingual},
and semantic role labeling \cite{kozhevnikov-titov-2013-bootstrapping,van-der-plas-etal-2011-scaling,akbik-etal-2015-generating}.
More directly relevant to this work,
word sense projection has been utilized to
generate sense-tagged corpora in target languages.
When combined with manual annotation,
these corpora frequently serve as ``gold'' standard test sets
for multilingual WSD systems,
such as those used in prior SemEval tasks 
\cite{moro-navigli-2015-semeval,navigli-etal-2013-semeval}.

A series of papers by \citet{oliver2012a,oliver2012b,oliver-climent-2014-automatic}
represents the first published work on using semantic projection
to associate words with lexical concepts.
Their approach transfers sense annotations from a source text to its translation
via word alignments.
The authors released the \emph{WN-Toolkit},
implementing their method.
However, due to the age of these tools, 
we were unable to obtain comparable results using them.

Most recently, \citet{martelli-etal-2023-lexicomatic} introduced \emph{LexicoMatic},
along with a manually-constructed benchmark dataset,
to algorithmically construct
``multilingual lexical-semantic dictionaries''.
This method also follows the projection framework,
but introduces a novel aggregation step.
For each source sense, it identifies the target-language words most frequently aligned to it,
and 
selects words 
until their cumulative frequency exceeds a threshold $\beta$.
These selected words 
are then added to the
corresponding synset,
thereby creating new senses in the target language.
Since their code and data are not available,
we compare our results to a re-implementation of their method.

While recent sense generation methods rely primarily on corpus statistics to filter alignments,
bilingual dictionaries have a history of improving automated word alignment.
Early work by \citet{probst-brown-2002-using} demonstrated how expanding bilingual dictionaries through statistical co-occurrence can 
enhance alignment quality.
Similarly, \citet{ren-etal-2007-improving} employed handcrafted dictionaries for the same purpose.
Building on this tradition, our work applies dictionary-based alignment and filtering
to the task of cross-lingual \task{},
replacing the corpus-aggregate thresholds of prior work
with knowledge-based resources.

Our work is motivated not only by the goal of validating a semantic theory,
but also by open challenges 
highlighted within
the semantics community.
\citet{neale2018} emphasizes the importance of wordnets
to semantic NLP and 
argues that the cost of 
manual construction motivates the 
development of automatic expansion techniques. \citet{mosolova2025} stress the continued importance
of research into creating word senses for low-resource languages,
and note that even large language models (LLMs) struggle with this task.
Recent work has explored using LLMs to generate natural language definitions 
for words in context, 
rather than assigning lemmas to pre-existing
concepts or synsets \cite{giulianelli-etal-2023-interpretable}.
While promising, this approach does not directly support wordnet expansion.
Similarly, \citet{teglia2025} argue that sense representations
in pretrained language models are sub-optimal and inefficient,
reinforcing the need for discrete, interpretable methods.
Taken together, these findings support the continued need for
explicit, knowledge-based systems
for cross-lingual
wordnet expansion.

\section{Methods}
\label{methods}

In this section, we describe the two principal components of 
our \task{} method: 
the sense projection pipeline, 
and the dictionary-based alignment algorithm.

\subsection{Sense Projection} \label{ourmethod}

Our method builds upon the semantic projection framework,
in which sense tags
from a source-language corpus are transferred 
to a target language through word alignment.
Because each sense tag corresponds to
a lexical concept,
projection associates target-language lemmas with 
existing sense tags, thereby generating new senses
in the target language.
This projection framework assumes a word-aligned parallel corpus
with sense annotations on the source side.
If a source word $x$ tagged with concept $s$
is aligned with a target word $y$,
then
the lemma of $y$ can also be associated with concept $s$.
We say that the semantic tag $s$ is \emph{projected} 
across the alignment link
from $x$ to $y$, which \emph{generates} the sense $(y, s)$.

Our method,  \exnet{}, 
introduces two dictionary-based enhancements to the projection framework.
First, it employs a novel aligner, \dbalign{}, which incorporates evidence from a bilingual dictionary
(Section \ref{dbalign}).
Second,
it applies a filtering step that discards projections lacking dictionary
support.
The full procedure is summarized in Figure~\ref{alg-bidictnet2025}.

The filtering step uses a bilingual dictionary to validate each aligned pair.
Let $D$ be a bilingual dictionary consisting of translation pairs $(x, y)$,
where $x$ and $y$ are words or phrases in the source and target languages that serve as valid translations of one another
in at least one context.
If a sense-tagged word $x$ is aligned with a target word $y$,
we project the sense only if $(x, y) \in D$.
Otherwise, the link is discarded and no sense is generated.
However, to facilitate the generation of senses involving rare words and proper nouns,
we always project a sense when $x$ and $y$ share the same orthographic form.

We optionally apply a part-of-speech (POS)
filter,
which discards projections when the
source and target words do not share the same
basic
POS (noun, verb, adjective or adverb).
Tags from standard 
tagsets (e.g., UPOS) are mapped to 
these four broad categories. 
Words falling outside these categories
are treated as function words and excluded from projection.

We expect our method to be particularly useful
in settings where comprehensive wordnets or sense inventories
are not yet available, 
including many low-resource languages.
While the semantic projection framework inherently assumes access to a bitext, 
we generate this automatically
by translating a
sense-annotated
source corpus into
the target language.
Our pipeline relies on machine translation,
a bilingual dictionary (for both alignment and filtering),
and an optional POS tagger.
Critically,
these resources are generally more accessible than sense inventories in many languages.

\begin{figure}[t]
{\small
\begin{algorithmic}
\Function{ExpandNet}{$\text{Bitext}$}
    \For{$(\text{sentence $S$, translation $T$)} \in \text{Bitext}$}
        \For{$(\text{content word $w$)} \in S$}
            \State $s \gets \text{Synset}(S, w)$
            \State $w_y \gets \text{\dbalign{}}(S, T, w)$
            \If{$(w, w_y)  \in Dict\;\lor\; w = w_y$}                 \State $s.\text{add}(w_y)$
        \EndIf
        \EndFor
    \EndFor
\EndFunction
\end{algorithmic}
\caption{Our \task{} algorithm, \exnet{}, where
$Dict$ is a set of pairs $(x, y)$ such that $y$ in the target language can express the same meaning as $x$ in the source language in some context,
$\dbalign{(S, T, w)}$ returns the word in $T$ aligned to $w$ in $S$, as described in Section~\ref{dbalign}, and
$Synset(S, w)$ returns the synset corresponding to the concept expressed by $w$ in $S$.
}
\label{alg-bidictnet2025}
}
\end{figure}

\subsection{Dictionary-Based Alignment}
\label{dbalign}

Dictionary-Based Align (\dbalign)
augments a pre-existing base aligner with 
evidence from a bilingual dictionary.
This integration improves the robustness of the base aligner's output
by prioritizing dictionary-supported alignments and explicitly handling multi-word
expressions (MWEs).

At a high level, \dbalign{}
chooses a set of alignment links
through a three-pass procedure.
First, it accepts
the intersection of links proposed by both the dictionary
and the base aligner. 
We treat these alignments as high-confidence,
as they are supported by both sources.
Next, it incorporates any remaining dictionary-suggested links
that do not conflict with previously accepted alignments. 
Finally, it adds remaining base-aligner links that 
do not conflict with the previously accepted set.
This ordering prioritizes high-confidence
alignments, before increasing coverage while
avoiding conflicts.
Throughout this process, two links are considered to \emph{conflict}
if they share a source or target token.
During each pass, \dbalign{} iterates over the source tokens 
and identifies target tokens that satisfy the current condition.
For example,
consider the first pass applied to the sentence pair
``\emph{that man is my husband}''
and 
``\emph{ese hombre es mi esposo}''.
Suppose the base aligner links
\emph{man} to
both \emph{hombre} and \emph{esposo}.
Because the bilingual dictionary contains the translation pair
(\emph{man}, \emph{hombre}) but not (\emph{man}, \emph{esposo}),
only the \emph{man}-\emph{hombre} link satisfies the first-pass condition and is added to the final set.
In subsequent passes,
any candidate links involving either
\emph{man} or \emph{hombre} will
conflict with this alignment and be discarded.

If a given token has multiple candidate links
that satisfy the condition of the current pass,
we resolve the tie using the \emph{diagonal heuristic}, 
adapted from the \texttt{fast\_align} system \cite{dyer-etal-2013-simple}.
The heuristic selects the link 
that minimizes the difference between the relative positions 
of the source and target tokens
within their respective sentences.
For example, consider aligning the sentences
\emph{``the taller man told the shorter one''}
and \emph{``el hombre más alto le dijo al hombre más bajo''}.
Suppose that two candidate links align
the English \emph{man} to both occurrences of \emph{hombre} in the target sentence.
The heuristic connects \emph{man} to the first \emph{hombre}, 
since their relative positions 
within their respective sentences are much closer.

To put it formally, the \dbalign{} algorithm proceeds as follows: 
Let 
$B$ be the set of alignment links proposed by the base aligner,
defined as all index pairs $(x, y)$ where the source token at index $x$
is aligned to the target token at index $y$.
Similarly, let $D$ be the set of
translation links supported by the dictionary, 
containing all index pairs $(x,y)$ 
where the source token at index $x$ is a translation
of the target token at index $y$.
This definition accommodates MWEs, provided they are present in the dictionary.
For example,
our English-Spanish dictionary maps the English MWE
\emph{fritter away}
to the Spanish \emph{derrochar}.
Because all MWEs in our dataset are pre-tokenized as single units,
\dbalign{} needs only to operate on individual tokens.
The final set of alignment links $A$ is then constructed from $B$ and $D$
using the three-pass algorithm described in detail in Figure~\ref{alg-dbalign}.

\begin{figure}[t]
{\small
\begin{algorithmic}
\Function{DBAlign}{$S, T, B, D$}
    \State $A \gets \emptyset$, $|S| \gets$ length of $S$, $|T| \gets$ length of $T$

    \For{$pass \in [ \textsc{Intersection}, \textsc{Dictionary}, \textsc{Base} ]$}

        \If{$pass = \textsc{Intersection}$}
            \State $C \gets B \cap D$
        \ElsIf{$pass = \textsc{Dictionary}$}
            \State $C \gets D$
        \Else
            \State $C \gets B$
        \EndIf

        \While{true}
            \State $\mathcal{V} \gets$ links in $C$ not incident on tokens already linked in $A$
            \If{$\mathcal{V} = \emptyset$}
                \State \textbf{break}
            \EndIf

            \State $(i,j) \gets$ \Call{DiagonalHeuristic}{$\mathcal{V}, |S|, |T|$}
            \State $A \gets A \cup \{(i,j)\}$
        \EndWhile

    \EndFor

    \State \Return $A$
\EndFunction
\\

\end{algorithmic}
\caption{Our dictionary-based alignment procedure, \textsc{DBAlign}. 
$B$ denotes links proposed by the base aligner, while $D$ denotes links between translations in the bilingual dictionary. $S$ and $T$ represent the source and target sentences, respectively.
Links are added in three passes, 
subject to a non-conflict constraint.
When multiple candidate links are available, the diagonal heuristic selects the link
closest to monotonic word order.}
\label{alg-dbalign}
}
\end{figure}

\section{Experiments}
\label{sec:experiments}

In this section, we describe our experimental setup,
resources, and evaluation framework.
We then introduce the baseline systems used for comparison.
Finally, we present results across four
language pairs,
including one used for development,
followed by a detailed error analysis and an ablation study
establishing the utility of the key innovations underlying \exnet{}.

\subsection{Setup}

\paragraph{\textbf{Corpus}}
Our source text is the SemCor English sense-annotated corpus \cite{miller1993},
specifically the XL-WSD version \cite{pasini2021xlwsd},
which is annotated with BabelNet synset IDs, streamlining cross-lingual evaluation.
We refer to this full, 37,176-sentence dataset as \corpusfull{}.
To expedite our preliminary experiments, 
we also constructed a 20\% sub-corpus, denoted as \corpustwenty{}, by sampling every fifth sentence.
\corpustwenty{} consists of 7,261 sentences containing 45,224 sense-annotated tokens.
These annotations cover 12,251 BabelNet synsets,
representing 9.7\%
of the noun and verb synsets
in Princeton WordNet.

\paragraph{\textbf{Translation}}
We utilize French as our development language and evaluate our method on
Spanish, Chinese, and Urdu.
For French, Chinese, and Spanish,
we translate the corpus using \texttt{gpt-4o-2024-11-20}
with the prompt 
``You are an expert translator. Translate from \{SOURCE\_LANGUAGE\} to \{TARGET\_LANGUAGE\}. Provide only the translation without explanations.''
These translations are tokenized, lemmatized, and POS-tagged using \texttt{spaCy}.
For Urdu, we 
obtain translations with \texttt{nllb-200-distilled-600M}\footnote{\url{https://huggingface.co/facebook/nllb-200-distilled-600M}}
and use
\texttt{LughaatNLP}\footnote{\url{https://github.com/MuhammadNoman76/LughaatNLP}} for the equivalent preprocessing.

\paragraph{\textbf{Alignment}}
As the base aligner for \dbalign{},
we employ SimAlign \cite{simalign},
configured to use byte-pair encoding (BPE) tokenization, 
embeddings from layer 8 of XLM-R,
and the \emph{itermax} matching algorithm.

\paragraph{\textbf{Dictionaries}}
For our English-Spanish and English-French dictionaries,
we take the union of all English-target translation pairs 
from WiktExtract \cite{ylonen2022} and PanLex \cite{panlex}.
From WiktExtract, we utilize all available pairs;
from PanLex, we obtain all translations of any word appearing in any Princeton WordNet synset via the NLTK interface.
For Chinese, 
we merge two freely available lexical resources: 
CC-CEDICT\footnote{\url{https://cc-cedict.org/wiki}} and ECDICT\footnote{\url{https://github.com/skywind3000/ECDICT}}. 
The former provides domain-general vocabulary, 
while the latter offers substantially broader coverage. We merge all translation pairs from both sources, 
removing duplicates.
Finally, for the English-Urdu dictionary, we combine 
dictionary files from GoldenDict\footnote{\url{http://goldendict.org/dictionaries.php}
from \url{https://sourceforge.net/projects/goldendict/}} 
with all English-Urdu pairs from WiktExtract. 
We then clean this concatenated dictionary 
by filtering out long phrases
and extraneous entries in
Hindi or Arabic.

\paragraph{\textbf{Additional Settings}} In the following experiments, we
employ the POS filter
in an effort to maximize precision.
However, the ablation study in Section~\ref{sec:ablation} reveals
that 
comparable results can be achieved without
using POS.

\subsection{Evaluation}

In this section, we describe our evaluation framework
for \task{}.
Given a candidate lemma-synset pairing,
the objective is to determine whether the proposed sense is valid,
that is, whether the lemma can lexicalize the corresponding concept.
Evaluating generated senses remains a challenging open problem \cite{neale2018}.

As an actively developed, well-established multilingual wordnet, 
BabelNet serves as the gold standard in our automatic evaluation.
Accordingly, a pairing of a word with a BabelNet synset by a \task{} system
is deemed correct if and only if
the synset contains the word.
For all experiments, we use the latest release, BabelNet 5.3.

We report the performance of 
sense generation methods using 
the following two metrics:
\begin{enumerate}
\item \emph{Sense Precision}: We first identify all proposed senses for which the
corresponding BabelNet synset contains at least one lemma in the target language.
We then calculate the proportion of these senses that are correct. 
\item 
\emph{Synset Coverage}:
We identify all
synsets attested in the source-side
corpus that contain at least one lemma in the target language.
We then calculate the proportion of these synsets
for which the method successfully proposes at least one correct sense.
\end{enumerate}

For both metrics, we restrict our evaluation to synsets containing at least one target-language lemma.
This restriction 
aims to account for incomplete coverage in BabelNet:
by definition, a lemma cannot appear in an empty synset, 
and any sense mapped to such a synset would therefore be deemed incorrect.
We argue that this formulation
is appropriate because the absence of lemmas in a synset for a given language 
is more likely to reflect a missing entry in BabelNet 
rather than a genuine lexical gap in 
the language itself. 
This distinction is crucial
for cross-lingual comparison,
as BabelNet coverage varies substantially across languages: 
within \corpustwenty{},
93.9\% of synsets contain at least one Spanish lemma,
compared to 83.7\% for French,
75.0\% for Chinese,
and only 17.5\% for Urdu.

We 
consider precision to be the most important measure of performance.
Compared to cross-lingual sense tagging, precision is vital for \task{}
because missed projections result in omissions that can be
mitigated by processing larger corpora,
making false positives the only source of true errors in the resulting output.
We do not report sense {\em recall} 
because there is no 
principled way to define a comprehensive ``ground set'' of all correct target-language senses
that could theoretically be projected given our corpus.
Such a ground set
cannot be derived from the sense-tagged source corpus, 
as a single source-language 
sense may be translated in multiple ways, including non-literally. 
Instead, {\em synset coverage} serves as a proxy for recall.
Its theoretical upper bound is reached 
when at least one correct sense is projected 
for every non-empty synset observed in the source corpus.
While no projection-based method is expected to achieve full coverage, 
this measure, unlike the raw number of correctly generated senses, enables meaningful comparison across different datasets and target languages.

\subsection{Comparison Systems}
\label{sec:comparison}

We evaluate \exnet{} against five comparison systems. 
Two are internal baselines: 
a dictionary-based approach and a direct LLM generation strategy.
The remaining three represent prior work, 
two of which are re-implementations of 
projection-based methods conceptually similar to \exnet{}.

First, we define a dictionary-centric baseline. 
This baseline naively assumes that, 
for a given sense in the source-side corpus, 
any target-language translation of the 
source word can express the corresponding meaning,
effectively ignoring all contextual 
information.

As a second baseline, we introduce a direct LLM generation approach.
We prompt an LLM to generate a target-language 
lexicalization for each concept
based on its definition.
This lexicalization is then paired with the concept to create a single sense.
Specifically, we use \texttt{gpt-5-chat-latest}
with the prompt ``You are a bilingual lexicon expert. Given a dictionary definition, produce the single word in \{LANGUAGE\} that best matches this definition. Provide only the \{LANGUAGE\} word without explanations!''
By design, this baseline produces exactly one sense per synset in the corpus. 
As our first prior-work baseline, we evaluate UWN.
Unlike the subsequent two baselines, 
UWN does not employ a projection framework.
Instead, it relies on a variety of resources,
including a small amount of manually labeled training data,
to train a supervised classifier capable of recognizing valid senses.
We obtain UWN in TSV format as provided by the authors.\footnote{\url{http://wordnets.org/}}
From this resource, 
we extract all lemma-synset pairs associated with a ``rel:means'' relation
where the lemma belongs to the target language.
We then map UWN's Princeton WordNet synset offsets to BabelNet synset IDs using the Python BabelNet API.\footnote{\url{https://babelnet.org/guide}}
For every synset in the corpus, 
we treat the corresponding target-language senses (lemma-synset pairs)
as the output of the method used for UWN.

Next, we re-implement the method from \citet{oliver-climent-2014-automatic},
which we refer to as \oconefour{}.
This approach uses two parameters, $i$ and $f$, 
to filter the generated senses.
Parameter $i$ specifies the minimum ratio between 
the highest and second-highest co-occurrence frequencies for any given synset.
In this context, a target word $y$ is said to ``co-occur'' with a synset $s$
if $y$ appears in a target sentence
whose source-side equivalent
contains a word
tagged with $s$.
Parameter $f$ dictates the maximum allowable ratio of the synset's frequency
to the word's frequency.
A candidate sense is projected 
only if its respective ratios 
satisfy both thresholds.
Following the original paper, we set $i = 2.5$ and $f = 5$.

We also re-implement the method of \citet{martelli-etal-2023-lexicomatic},
the current state of the art for sense generation,
which we refer to as \mtwothree{}.
The method uses a single parameter, $\beta$, to threshold the total frequency
of the target words retrieved for each source sense.
The authors report results for $\beta=0.7$, $\beta=0.9$, and $\beta=1.0$.
We use $\beta=0.7$ in our experiments, as this setting 
achieved the best performance among the three.

In our re-implementations of these
pre-existing methods, we control
as many variables as possible. \oconefour{} and \mtwothree{} originally
used their own translations, one 
manual and one automatic, to project senses.
In our experiments, we instead use the same
translations employed by our main \exnet{} method.
Similarly, rather than the in-house aligner described 
for \mtwothree{},
we use SimAlign,
the base aligner of \dbalign{}.

Both \oconefour{} and \mtwothree{}
make use of corpus-level statistics when deciding whether to project a sense annotation.
In our experiments,
we apply them to the same corpora as \exnet{},
which may differ from their original setup.
While this may affect their performance, 
we include them to provide a consistent comparison across methods 
and to highlight 
that the performance of \exnet{} is expected to be invariant to corpus size.

\subsection{Results}
\label{sec:results}

\begin{table*}[t]
\tablesize
\centering
\begin{tabular}{l|ccc|ccc|ccc|ccc}
\toprule
 &
\multicolumn{3}{c|}{\textbf{French (dev)}} &
\multicolumn{3}{c|}{\textbf{Spanish}} &
\multicolumn{3}{c}{\textbf{Chinese}} &
\multicolumn{3}{c}{\textbf{Urdu}} \\
\cmidrule(lr){2-4} \cmidrule(lr){5-7} \cmidrule(lr){8-10} \cmidrule(l){11-13}
Method & Prec. & Cov. & HM & Prec. & Cov. &HM & Prec. & Cov. &HM & Prec. & Cov. &HM \\
\midrule
\uwn        & 46.1 & 56.9 & 50.9 & 71.9 & 51.4& 60.0 & 45.0 &  5.8 & 10.3 & \textbf{59.9} & 13.1 & 21.5 \\
\dictbaseline & 5.2 & \textbf{93.7} & 9.9 & 8.5 & \textbf{88.2} & 15.5 & 7.8 & \textbf{75.1} & 14.1 &  9.4 & 44.0 & 15.5  \\
\oconefour   & 71.1 &  8.2 & 14.6 & 72.0 & 9.9 & 17.4 & 34.3 & 2.8 & 5.2 &  44.0 & 0.5 & 1.0 \\
\mtwothree  & 52.3 & 67.2 & 59.0 & 55.3 & 67.8 & 60.9 & 34.7 & 48.4 & 40.6 & 6.3 &  24.4 & 10.0 \\
\llmbase       & 44.3 & 44.3 & 44.3 & 55.3 & 55.3 & 55.3  & 41.8 & 41.8 & 41.8 &  50.5 & \textbf{50.5} & \textbf{50.5} \\
\exnet      & \textbf{71.5} & 69.7 & \textbf{70.6} &  \textbf{82.8} & 66.6 & \textbf{73.8} & \textbf{64.8} & 44.4 & \textbf{52.7} &  56.1 &  28.6 & 37.9 \\
\bottomrule
\end{tabular}
\caption{Sense Precision (Prec.), Synset Coverage (Cov.), and harmonic mean of the two (HM) in percentages for \exnet{} and comparison methods,
including the dictionary baseline (\dictbaseline{}), 
LLM baseline (\llmbase{}),
two prior work re-implementations (\oconefour{}, \mtwothree{}), and UWN.
Results are reported on the \corpustwenty{} corpus, 
across four target languages: French (our development language), Spanish, Chinese, and Urdu.
The highest number in each column is bolded.
}

\label{tab:results_sc20}
\end{table*}

Table \ref{tab:results_sc20} summarizes the results of our experiments on \corpustwenty{}. 
While the dictionary baseline achieves 
high coverage for most languages,
its precision is low enough to make it impractical for many use cases.
This highlights the importance of context-aware 
translation and reinforces why bitexts are 
typically used for this task.
The LLM baseline achieves fairly consistent performance across languages.
Because it was prompted to provide only one sense per synset,
its coverage could likely be increased by allowing multiple senses.
However, there is no reason to expect this would improve its precision, which is below that of
\exnet{} in all languages. 
These results suggest 
that \task{} remains challenging 
for LLMs.
Nevertheless, an LLM-based approach could
serve as a useful back-off for senses 
that are not found in the projected corpus, 
such as rare senses or those corresponding to lexical gaps in the source language.

We now turn to the performance of methods
from prior work.
\uwn{} 
achieves competitive coverage for French and Spanish,
but its precision varies across languages.
Coverage in Urdu and Chinese is poor,
although \uwn{} is more precise
than \exnet{} in Urdu---the 
only case in which any system surpasses
\exnet{}
in precision.
\oconefour{} 
attains relatively high precision on French and Spanish,
but exhibits the lowest coverage across all languages, 
making it ill-suited for populating a wordnet. The most recent comparison method,
\mtwothree{},
achieves substantially better coverage,  
but at the cost of lower precision overall. 
Table~\ref{tab:results_scfull} summarizes the results of our experiment on \corpusfull{}. 
The trends are broadly similar, but some differences can be observed.
For one, \oconefour{}'s precision improves slightly on French, Spanish and Urdu, 
placing it above that of \exnet{} in those languages.
The precision of \uwn{} improves slightly for Chinese, putting it above that of \exnet{}.
Additionally, \llmbase{} now outperforms \exnet{} with respect to Chinese coverage.

The decrease in \exnet{}'s precision from the \corpustwenty{} split to the full \corpusfull{} dataset may seem surprising;
in fact, it is actually a result of how the evaluation is measured rather than a drop in performance.
Under our standard evaluation framework, which counts only distinct generated senses, \exnet{} achieves an average precision of 82.7\% across the individual 20\% splits of \corpusfull{}.
This precision drops to 75.8\% on the full dataset.
However, when evaluating every generated sense including repetitions, precision remains consistently around 90\% for the splits and the full dataset.
This shows that any given sense the model produces is highly likely to be correct.

The difference is due to frequency.
\exnet{} reliably generates common senses correctly, often multiple times.
However, as a dataset grows, the model quickly finds most common correct senses, 
while rare, unique errors continue to add up over time.
As a result, a larger dataset simply collects more unique errors, 
which lowers precision when evaluation is restricted to distinct generated senses.

Urdu stands as an outlier on both \corpustwenty{} and \corpusfull{}.
Many systems, particularly \exnet{}, perform noticeably worse on Urdu than on the other studied languages.
While this decline is visible in Sense Precision, \exnet{} especially struggles with Synset Coverage.
Because \exnet{} relies heavily on translation, alignment, and dictionaries, its performance is expected to suffer in a lower-resource setting.
This resource disparity is visible in the employed resources; 
for example, our English-Urdu dictionary contains translations for only 22,855 source-side words,
compared to 98,377 in our English-Spanish dictionary.

Beyond basic resource scarcity, we identified specific issues with translation and gold-standard data quality that further limit performance.
For translation into Urdu, we utilized \texttt{NLLB} due to its strong baseline for low-resource languages.
However, a targeted error analysis of cases where \exnet{} failed due to translation errors (while \uwn{} succeeded) showed that \texttt{GPT} produced subjectively better translations in 17 out of 26 instances.
Finally, while we have already discussed BabelNet's decreased coverage in Urdu, this drop in performance may also be shaped by the poor \emph{quality} of said coverage.
Manual inspection revealed numerous erroneous inclusions, such as a Hindi word transliterated into the Perso-Arabic script (``\includegraphics[height=1em]{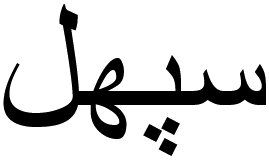}'').
More broadly, 6.9\% of all Urdu lemmas present in the BabelNet concepts within \corpusfull{} appeared to be English, containing only Latin characters and lacking Perso-Arabic text entirely.

Overall, \exnet{} achieves strong performance across languages and metrics.
It typically proves more precise than 
methods from prior work on modestly sized corpora,
and it consistently outperforms them in the harmonic mean of precision and coverage on all evaluated corpora.
Using the \corpusfull{} corpus, it generates 18,723 correct senses in Spanish,
16,046 in French, 9,841 in Chinese, and 971 in Urdu.
These 
results highlight the effectiveness of \exnet{}'s
sentence-level approach,
which operates
independently 
of corpus-level statistics,
making it robust to variations
in corpus size.

\begin{table*}[t]
\tablesize
\centering
\begin{tabular}{l|ccc|ccc|ccc|ccc}
\toprule
 &
\multicolumn{3}{c|}{\textbf{French (dev)}} &
\multicolumn{3}{c|}{\textbf{Spanish}} &
\multicolumn{3}{c}{\textbf{Chinese}} &
\multicolumn{3}{c}{\textbf{Urdu}} \\
\cmidrule(lr){2-4} \cmidrule(lr){5-7} \cmidrule(lr){8-10} \cmidrule(l){11-13}
Method & Prec. & Cov. & HM & Prec. & Cov. &HM & Prec. & Cov. &HM & Prec. & Cov. &HM \\
\midrule
\uwn        & 48.1 & 53.8 & 50.8 & 72.8 & 48.6 & 58.3 & \textbf{61.4} & 6.3 &  11.4 & \textbf{59.9} & 7.5 & 13.3 \\
\dictbaseline & 5.5 & \textbf{90.8} & 10.5 & 8.5 & \textbf{84.1} & 15.4 & 8.0 & \textbf{77.0} & 14.6 &  8.7 & 36.7 & 14.0  \\
\oconefour   & \textbf{74.0} & 9.4  & 16.7 & \textbf{76.0} & 11.2 & 19.5 & 32.7 & 2.6 &  4.8 & 55.3  & 0.6 & 1.2 \\
\mtwothree  & 43.2 & 67.4 & 52.7 & 45.7 & {67.0} & 54.3 & 26.4 & 47.3 & 33.9 & 23.2 & 27.2 & 25.0 \\
\llmbase       & 56.7 & 56.7 & 56.7 & 59.2 & 59.2 & 59.2  & 51.1 & {51.1} & \textbf{51.1} & 45.7  & \textbf{45.7} & \textbf{45.7} \\
\exnet      & 63.4 & {67.9} & \textbf{65.6} & 75.8  & 63.2 & \textbf{68.9} & 56.4 & 43.0 & 48.8 &  46.6 & 24.6  & 32.2 \\
\bottomrule
\end{tabular}
\caption{Sense Precision (Prec.), Synset Coverage (Cov.), and harmonic mean of the two (HM) in percentages for \exnet{} and comparison methods,
including the dictionary baseline (\dictbaseline{}), 
LLM baseline (\llmbase{}),
two prior work re-implementations (\oconefour{}, \mtwothree{}), and UWN.
Results are reported on the \corpusfull{} corpus, 
across four target languages: French (our development language), Spanish, Chinese, and Urdu.
The highest number in each column is bolded.
}

\label{tab:results_scfull}
\end{table*}

\subsection{Error Analysis}
\label{sec:error_analysis}

While \exnet{} performs well relative to prior work, 
absolute scores remain moderate, 
particularly for Chinese and Urdu. 
This may actually reflect constraints in the evaluation setup, 
as a projection is counted as correct only if the projected sense appears in BabelNet, 
whose coverage varies substantially across languages. 
We restrict evaluation to synsets that are not empty. 
However, 
even these may lack many valid lemmas, 
especially in languages that are typologically distant from English or less densely represented in multilingual lexical resources. 
Consequently, automatic evaluation may incorrectly label valid projections as errors. 
In such cases, the system may in fact be proposing legitimate lexicalizations that are absent from the resource, 
suggesting that \exnet{} can serve not only as a tool for extending existing multilingual lexicons,
but also as an evaluation target. 

To investigate this issue more carefully, 
we conducted a targeted manual analysis. 
We focused on Chinese and randomly sampled 20 projected senses that were marked incorrect due to their absence from BabelNet. 
Each case was reviewed with the assistance of a native speaker, 
who judged whether the projected lemma appropriately expressed the intended concept in context.

Our analysis suggests that many apparent errors
produced by \exnet{}
are, in fact, valid projections that fall outside the coverage of BabelNet.
Of the 20 sampled Chinese 
senses marked incorrect, 
14 were judged to be valid.
For example, the
concept ``\emph{by contrast; on the other hand}''
was projected from the 
English word \emph{however}
to the Chinese word \begin{CJK}{UTF8}{gbsn}然而\end{CJK}.  
This usage is valid, but missing from BabelNet,
so it was counted as incorrect.

The remaining cases primarily reflect noise 
in upstream tools and resources.
Three errors arose from inaccurate source-side sense annotations.
For example,
in the phrase ``\emph{eyes which had paled with the years}'',
the word ``\emph{years}''
is tagged with the sense
``\emph{a late time of life}'',
but in context refers to literal time passing.
The aligned Chinese term \begin{CJK}{UTF8}{gbsn}岁月\end{CJK}
is appropriate given the English source, 
but the annotation itself 
does not match the intended meaning.
Two errors resulted from 
dictionary issues.
For instance, ``\emph{alizarin crimson}'' was aligned with \begin{CJK}{UTF8}{gbsn}茜\end{CJK} (``\textit{crimson}''). 
This occurs because the bilingual dictionary contains an entry linking ``\emph{alizarin crimson}'' to \begin{CJK}{UTF8}{gbsn}茜\end{CJK}, 
even though the correct translation is the phrase \begin{CJK}{UTF8}{gbsn}茜红色\end{CJK}. 
As a result, 
the projection incorrectly passes the dictionary filter. 
A similar issue occurs in another case involving the word ``\textit{date}'', 
where the dictionary contains the pair of ``\textit{date}'' and \begin{CJK}{UTF8}{gbsn}约会\end{CJK}, 
but not the appropriate \begin{CJK}{UTF8}{gbsn}约会对象\end{CJK} (``\textit{dating partner}'') in context.
The final analyzed case reflects a translation error: 
in a financial context, 
\begin{CJK}{UTF8}{gbsn}冒险\end{CJK} was used to translate the English word \emph{venture}, 
though \begin{CJK}{UTF8}{gbsn}冒险\end{CJK} more commonly denotes literal risk-taking 
and does not fully capture the intended business meaning.

The native speaker also analyzed cases involving sense-tagged tokens that were not aligned and thus not projected.
While the incorrect projections discussed above negatively impact precision, 
these unaligned tokens represent errors that decrease coverage.
We analyzed all such instances within the first 200 lemmas of the 
\corpustwenty{} dataset, identifying 10 unaligned lemmas in total.
Of these 10, 5 are found to be cases of nonliteral translations,
where the sentence translation, while correct, does not have a direct equivalent for the source
word in question.
For example, in the Chinese phrase \begin{CJK}{UTF8}{gbsn}``已经多久了''\end{CJK}, 
a translation of the English ``\emph{How long has it been...}'',
there is no direct equivalent to the word \emph{been}, 
even though the translation is correct.
In the remaining 5 cases, the system failed to align an MWE present in the English source, even when a valid translation existed.
For example, in the sentence ``\emph{Is your purchasing agent offering...}'',
the alignment system fails to link the MWE ``purchasing agent'' to its valid Chinese translation,
\begin{CJK}{UTF8}{gbsn}``采购代理''\end{CJK},
present in the target sentence.
This is due, in part, to the fact that the Chinese term \begin{CJK}{UTF8}{gbsn}``采购代理''\end{CJK} is not found in the dictionary employed by our aligner.

These findings are consistent with observations across other languages.
Analysis of Urdu
suggests a similar or higher
proportion of 
``false alarm'' errors
arising from BabelNet omissions.
While we did not have native speakers for Spanish or French, 
inspection of flagged errors
indicates 
that most arose either
from similar BabelNet omissions or from noise in
tools or resources, rather than methodological failure.

In summary, many errors identified
by automatic evaluation
are attributable to gaps in BabelNet rather than incorrect projections.
Remaining errors are primarily caused by mistakes in annotation, 
alignment, or translation resources, rather
than by the method itself.

\subsection{Ablation}
\label{sec:ablation}

In this section, we present the results of an ablation study,
intended to quantify the impact of the principal innovations of \exnet{}.
The target language for these experiments is Spanish.
We test four additional 
variants of our method.
One ablative axis is the use of our dictionary-based aligner, \dbalign{};
we consider instead using the base aligner, SimAlign, on its own.
Additional axes
include the use of our dictionary and
POS filters.
Because \dbalign{} is inherently dictionary-based, 
employing it while omitting the dictionary filter yielded minimal differences;
therefore, we do not report this setting.

Table \ref{tab:ablation} shows the results of these experiments.
The results clearly indicate that both of our key innovations, \dbalign{} and
the dictionary-based filter, have quantifiable benefits:
The dictionary filter  dramatically improves precision, from 49.1\% to 83.9\%,
at low cost in terms of coverage.
Using \dbalign{} increases coverage significantly.
Using our two main innovations together 
yields the best overall balance, with 
no major sacrifices.
We therefore conclude that:
(1) Our dictionary filter improves precision.
(2) Our dictionary-based aligner improves coverage.
(3) Use of both produces the best overall result.

Importantly, our system performs well even 
without the POS filter.
The system which uses \dbalign{}, the dictionary filter, and no POS filter achieves the best harmonic mean
of precision and coverage.
Overall, this indicates that the POS filter is not essential for 
strong performance, though we choose to use it due to our desired emphasis on precision.
The usefulness of such a filter may vary across languages and use cases. \begin{table}[t]
\tablesize
    \centering
    \begin{tabular}{c|c|c|c|c|c}
    \toprule
        Aligner     &  Dict. & POS &  Prec. & Cov.  & HM  \\
        \midrule
        \dbalign{}    &  Used   & Used &  82.8 &  66.6 & 73.8  \\
        \dbalign{}    &  Used    & Off  & 80.8 &  68.3 & 74.0  \\
                                        SimAlign    &  Used  & Used  & 84.1 & 62.2 & 71.5 \\
        SimAlign    &  Used  & Off   & 83.9 & 59.5 & 69.6 \\
                SimAlign    &  Off & Off   & 49.1 & 68.1 & 57.1 \\
        
                                \bottomrule
    \end{tabular}
    \caption{Sense Precision (Prec.), Synset Coverage (Cov.), and harmonic mean of the two (HM) in percent
    for five variants of \exnet{} on \corpustwenty{},
    with Spanish as the target language.
    ``Dict.'' indicates whether a dictionary-based filter was used, while ``POS'' indicates whether a POS filter was used.}
    \label{tab:ablation}
\end{table}

\section{Conclusion}
\label{conclusion}

\exnet{} establishes a competitive standard for \task{},
delivering strong results across four language pairs
and demonstrating scalability to small corpora.
Our method achieves higher precision than previously published systems,
without relying on corpus-level statistics
to filter spurious alignments.
Instead, it uses an innovative dictionary filter, which,
together with \dbalign{}, our dictionary-based aligner, 
is shown to improve \task{} performance.
Human analysis indicates that most errors are not inherent to the method,
but rather arise from gaps in resources
and noisy tools.
Overall, our work represents a step forward for \task{} and its applications,
providing a solid foundation for further developments.

\section*{Limitations}

A primary limitation of \exnet{} is its reliance on pre-existing linguistic resources.
Consequently, its performance and applicability may vary significantly
depending on a language's available data.
Additionally, translation and projection may not necessarily be sufficient, 
on their own, to create and populate multilingual wordnets.
This is particularly evident in cases of lexical gaps, concepts which lack a single-word lexicalization in
a given language, necessitating the use of a compositional phrase.
It is unlikely that projection would allow for the creation of senses
in a target language that correspond to lexical gaps in the source language.
Finally, our optimal results currently rely on
proprietary GPT models.
While the \exnet{} framework is theoretically compatible with alternative
translation tools, including open-source models,
GPT provided the best performance in our experiments.
As such, replicating the exact results presented in this paper
relies on closed-source architecture
and incurs financial costs associated with API usage.

\section*{Acknowledgements}

We thank Sevryn Robinson and Junhyeon Cho for their assistance in refining our algorithm and applying it to Korean.
This research was supported by the Natural Sciences and Engineering Research Council of Canada (NSERC) and the Alberta Machine Intelligence Institute (Amii).

\printbibliography[heading=subbibintoc]

\end{document}